\documentclass[lettersize,journal]{IEEEtran}
\usepackage{amsmath,amsfonts}
\usepackage{array}
\usepackage[caption=false,font=normalsize,labelfont=sf,textfont=sf]{subfig}
\usepackage{textcomp}
\usepackage{stfloats}
\usepackage{url}
\usepackage{verbatim}
\usepackage{graphicx}
\usepackage{cite}

\usepackage[breaklinks,colorlinks,citecolor=tkdeblue]{hyperref}

\usepackage{booktabs}
\usepackage{color}
\usepackage{bbding} 
\usepackage{algorithm}
\usepackage{algpseudocode}
\usepackage{multirow}
\usepackage[none]{hyphenat}
\usepackage[table,xcdraw,svgnames,dvipsnames]{xcolor}

\definecolor{cred}{HTML}{FF6B6B}
\definecolor{cyellow}{HTML}{FEC260}
\definecolor{cgreen}{HTML}{70AD47}
\definecolor{cblue}{HTML}{4D96FF}
\definecolor{cpurple}{HTML}{2A0944}
\definecolor{ggray}{RGB}{127,127,127}
\definecolor{aliceblue}{rgb}{0.54, 0.77, 1.0}
\definecolor{tkdeblue}{rgb}{0.21,0.49,0.74}

\def\BibTeX{{\rm B\kern-.05em{\sc i\kern-.025em b}\kern-.08em
    T\kern-.1667em\lower.7ex\hbox{E}\kern-.125emX}}
\usepackage{balance}

\begin{document}

\title{From Query to Explanation: \\Uni-RAG for Multi-Modal Retrieval-Augmented Learning in STEM}
\author{Xinyi Wu, Yanhao Jia, Luwei Xiao, Shuai Zhao, Fengkuang Chiang, Erik Cambria
\thanks{This paper was partially presented at the 63rd Annual Meeting of the Association for Computational Linguistics (ACL2025), paper no. 2025.main-acl.1171.}
\thanks{Xinyi Wu and Fengkuang Chiang are with the School of Education, Shanghai Jiao Tong University, Shanghai, China, 200240 (E-mail: summer.xywu, fkchiang@sjtu.edu.cn).}
\thanks{Yanhao Jia, Shuai Zhao and Erik Cambria are with the College of Computing and Data Science, Nanyang Technological University, Singapore, 639798 (E-mail: yanhao002@e.ntu.edu.sg, shuai.zhao, cambria@ntu.edu.sg).}
\thanks{Luwei Xiao is with the school of Computer Science and Technology, East China Normal University, Shanghai 200062 (E-mail: louisshaw@stu.ecnu.edu.cn).}
}

\markboth{Preprint}%
{From Query to Explanation: Uni-RAG for Multi-Modal Retrieval-Augmented Learning in STEM}

\maketitle

\begin{abstract}

In AI-facilitated teaching, leveraging various query styles to interpret abstract educational content is crucial for delivering effective and accessible learning experiences. However, existing retrieval systems predominantly focus on natural text-image matching and lack the capacity to address the diversity and ambiguity inherent in real-world educational scenarios.
To address this limitation, we develop a lightweight and efficient multi-modal retrieval module, named Uni-Retrieval, which extracts query-style prototypes and dynamically matches them with tokens from a continually updated Prompt Bank.
This Prompt Bank encodes and stores domain-specific knowledge by leveraging a Mixture-of-Expert Low-Rank Adaptation (MoE-LoRA) module and can be adapted to enhance Uni-Retrieval’s capability to accommodate unseen query types at test time.
To enable natural language educational content generation, we integrate the original Uni-Retrieval with a compact instruction-tuned language model, forming a complete retrieval-augmented generation pipeline named Uni-RAG.
Given a style-conditioned query, Uni-RAG first retrieves relevant educational materials and then generates human-readable explanations, feedback, or instructional content aligned with the learning objective.
Experimental results on SER and other multi-modal benchmarks show that Uni-RAG outperforms baseline retrieval and RAG systems in both retrieval accuracy and generation quality, while maintaining low computational cost.
Our framework provides a scalable, pedagogically grounded solution for intelligent educational systems, bridging retrieval and generation to support personalized, explainable, and efficient learning assistance across diverse STEM scenarios.
\end{abstract}

\section{Introduction}

\IEEEPARstart{A}{rtificial} Intelligence for Education (AI4EDU) has emerged as a transformative force, harnessing advanced AI techniques to enhance instructional design, learning processes, and assessment across diverse educational contexts, demonstrating tremendous potential in various educational scenarios \cite{hwang2020vision,wang2023improving}.
For example, education-related question-answering systems and multi-modal retrieval systems built upon large language models (LLMs)~\cite{yang2025qwen3,prompt14} have achieved state-of-the-art performance.
However, beneath this prosperity lie potential challenges.
In particular, the global emphasis on Science, Technology, Engineering, and Mathematics (STEM) education has led to an explosion of interdisciplinary resources—textual explanations, diagrams, interactive simulations, and multi-media content—that traditional retrieval systems struggle to accurately present to educators and learners \cite{intro3}.

Current retrieval frameworks are primarily optimized for natural language and high-resolution image matching, making them well-suited for conventional text–image tasks. However, they lack the flexibility to accommodate diverse query modalities, such as low-resolution sketches, audio descriptions, and domain-specific educational diagrams \cite{li2024alleviating}. In particular, their inability to handle noisy or imprecise inputs—such as fuzzy visuals or audio signals—often leads to degraded retrieval accuracy. These limitations result in imprecise or biased outcomes, ultimately hindering access to pedagogically valuable resources and restricting the potential of AI4EDU applications.
With the growing demands of AI in education, it is crucial to develop retrieval algorithms that offer enhanced modality adaptability and semantic generalization.
Furthermore, educational scenarios are often sensitive to inference costs, making it essential to strike a balance between performance and computational efficiency in next-generation retrieval systems.

Despite advancements in text-image matching techniques \cite{intro12,intro11}, current retrieval systems still encounter challenges when implemented in STEM education \cite{li2025freestyleret}.
These models are primarily optimized for matching text and images, neglecting the variety of query types essential in educational scenarios, including voice, sketches, and low-resolution images \cite{intro6}.
To overcome these shortcomings, we introduce a multi-style, multi-modal retrieval architecture tailored to STEM education scenarios, named \textbf{Uni-Retrieval}. The Uni-Retrieval includes a novel plug-and-play feature representation structure called Prompt Bank, which provides a more universal representation of information across different disciplines and semantic contexts by matching the abstract features of the data. By extracting query-style prototypes and dynamically retrieving style-relevant prompt tokens from the continuously updated Prompt Bank, Uni-Retrieval unifies visual and textual inputs into a shared representation space. This enables robust and versatile retrieval across heterogeneous query formats, all without the need for full-parameter fine-tuning of large-scale foundation models.

Building upon this efficient retrieval backbone, we propose \textbf{Uni-RAG}, the first end-to-end Retrieval-Augmented Generation system tailored for STEM domains, which seamlessly integrates Uni-Retrieval with the Qwen3 model, as illustrated in Fig.~\ref{fig:motivation}.
From query to explanation, Uni-RAG bridges retrieval and generation by first retrieving relevant educational materials based on a style-conditioned query, and then generating human-readable explanations, feedback, or instructional content aligned with the underlying learning objective.
Specifically, Uni-RAG (1) encodes user queries with style-aware prototypes, (2) retrieves the top-$k$ semantically aligned document chunks from a STEM-focused knowledge base, and (3) conditions Qwen3 model on these retrieved contexts to generate precise, contextually grounded answers. 
To validate the effectiveness of the proposed algorithm, we conducted a series of experiments using the SER dataset as the primary benchmark, supplemented by four additional retrieval datasets for comprehensive comparison. Compared to baseline models, our approach consistently achieves superior performance, demonstrating strong generalization capabilities while maintaining high computational efficiency.
Our main contributions are summarized as follows\footnote{This work is an extension of our preliminary findings, which have been accepted for presentation at the conference of ACL 2025. Please see reference \cite{jia2025uni} for the conference version of this paper. The current submission introduces a novel framework designed to offer robust retrieval and explanation tools for the educational community.}:

\begin{figure*}[!t]
    \centering
    \includegraphics[width=\linewidth]{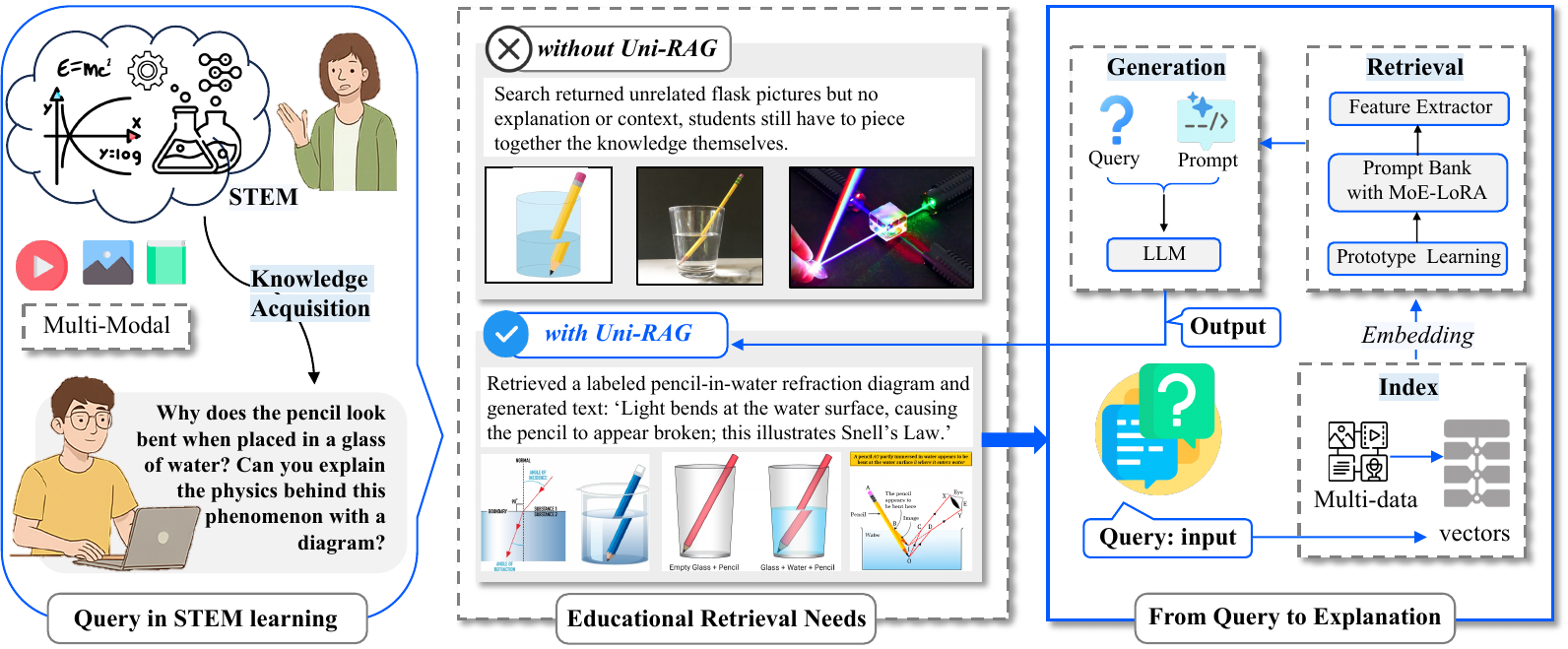}
    \vspace{-5mm}
    \caption{This advancement provides a scalable and precise solution for diverse educational needs. Previous retrieval models focus on text-query retrieval data or simple image-text retrieval. Our style-diversified retrieval setting accommodates the various query styles preferred by real educational content.}
    \label{fig:motivation}
    \vspace{-5mm}
\end{figure*}

\begin{itemize}
    \item Multi-Style Retrieval-Generation Integration: we demonstrate how prototype-guided prompt tuning can drive both retrieval precision and downstream generative relevance in a unified RAG pipeline. To the best of our knowledge, the proposed method is the first study to enhance retrieval through a unified RAG pipeline in the STEM domain.
    
    \item Efficient Prompt Bank Extension: we extend the Prompt-Bank to MoE-LoRA Prompt Bank, which  allows the retriever to specialize along multiple stylistic axes—visual, textual, sketch-based, and audio—while keeping the additional parameter footprint minimal. Through sparse expert activation, we preserve inference speed and maintain a lean memory profile, making the system practical for real-time educational applications.
    
    \item Empirical Validation: through comprehensive evaluations on STEM educational queries, we show that Uni-RAG outperforms baseline RAG systems in retrieval accuracy, generation quality, and end-user satisfaction. We hope that our method contributes to advancing the development of the AI4EDU community.
\end{itemize}

\section{Related Works}
\label{sec:related_work}

\subsection{Multi-task Learning}
In STEM education, the multi-style retrieval model needs to leverage multi-task learning to align features and learning across different modal samples. Multi-task learning refers to the simultaneous training and optimization of multiple related tasks within a single model \cite{9392366,xiao2025exploring,jia2025seeing,jia2025robust}. By sharing parameters and representations across functions, it improves overall performance. Compared to other transfer learning methods, including domain adaptation \cite{farahani2021brief} and domain generalization \cite{zhou2022domain}, multi-task learning annotates data and achieves CLIP-level model fine-tuning and convergence, the data of each task in multi-task learning is well-labeled. 

Overall, multi-task learning introduces a new tool for STEM education practitioners that may help meet requirements, especially if speed and efficiency are preferred over performance. While many recent multi-task learning employ two clusters of contemporary techniques, hard parameter sharing and soft parameter sharing \cite{ruder2017overview}. In hard parameter sharing, most or all of the parameters in the network are shared among all tasks \cite{kokkinos2017ubernet}. In soft parameter sharing,the models are tied together either by information sharing or by requiring parameters to be similar \cite{yang2016trace}. Consequently, our Uni-Retrieval adopts a blended multi-task learning paradigm, adopt both hard and soft  parameter in different styles of tasks. Building upon successul multi-task learning method for CLIP, such as CoCoOP \cite{zhou2022conditional}, MaPLe \cite{khattak2023maple}, and FreestyleRet \cite{li2025freestyleret}, our study leverages these techniques to strengthen domain adaptation and multi-task learning. 

\subsection{Query-based Retrieval}
Existing work in Query-based Image Retrieval (QBIR) primarily includes content-based image retrieval \cite{chen2022deep}, text-based image retrieval \cite{li2011text}, and multi-modal retrieval \cite{neculai2022probabilistic}. In content-based image retrieval, the visual features of images are directly utilized for retrieval. However, its reliance on fixed content and location makes it relatively inflexible in capturing diverse user intents \cite{lee-etal-2024-interactive}. Alternative methods like sketching \cite{chowdhury2022fs, chowdhury2023scenetrilogy} and scene graph construction \cite{johnson2015image} enable the retrieval of abstract images that are hard to describe verbally, though they lack the intuitive ease of natural language-based retrieval. In text-based image retrieval, enhancements to text queries often involve indicating content structure.
Despite advancements in QBIR, challenges including the semantic gap that can lead to inaccurate retrieval results, high computational complexity and resource costs for large-scale image databases, and the high cost of obtaining quality data annotations \cite{li-etal-2024-generative}. The application of QBIR to educational resource retrieval is promising but has been hindered by the complexity of educational discourse, the limitations of educational databases, and the associated costs \cite{zhou-etal-2024-vista}.  Our query model effectively combines multi-modal retrieval methods, integrating audio and natural language with multi-style image inputs. The former enables natural and rapid expression of content, while the latter facilitates accurate and intuitive image localization, enhancing educational data retrieval.

\subsection{Prompt Tuning}
Prompt tuning \cite{brown2020language} was first proposed in natural language processing~(NLP) and has been an efficient approach that bridges the gap between pre-trained language models and downstream tasks \cite{li2023weakly,zhao2023prompt,zhao2024unlearning}. Prompt tuning leverages natural language prompts to optimize the language model’s ability to understand tasks, which demonstrates exceptional performance in few-shot and zero-shot learning. 
Recent studies have focused on optimizing various components of prompt tuning, such as  prompt generation, continuous prompt optimization \cite{prompt3}, and adapting to large-scale models through methods like in-context learning~\cite{dong2024survey,zhao2024universal}, instruction-tuning~\cite{wang2024pandalm}, and chain-of-thought~\cite{prompt4}. 
For example, Lester et al. \cite{lester2021power} leverage soft prompts to condition frozen language models to enhance the performance of specific downstream tasks. 
Long et al. \cite{long2024prompt} propose an adversarial in-context learning algorithm, which leverages adversarial learning to optimize task-related prompts.
Furthermore, prompt tuning has gradually become a pivotal technique in computer vision~\cite{shen2024multitask}, enabling efficient adaptation of pre-trained models to diverse tasks. Notable methods include visual prompt tuning for classification~\cite{jia2022visual}, learning to prompt for continual learning~\cite{wang2022learning}, context optimization and conditional prompt learning for multi-modal models~\cite{zhou2022conditional}, and prompt-based domain adaptation strategies~\cite{ge2023domain}.
For example, Nie et al. \cite{nie2023pro} introduce the pro-tuning algorithm for learning task-specific vision prompts, applied to downstream task input images with the pre-trained model remaining frozen.
Shen et al. \cite{shen2024multitask} leverage cross-task knowledge to optimize prompts, thereby enhancing the performance of vision-language models and avoiding the need to independently learn prompt vectors for each task from scratch.
Cho et al. \cite{cho2023distribution} introduce distribution-aware prompt tuning for vision-language models, optimizing prompts by balancing inter-class dispersion and intra-class similarity.
Despite significant advancements in previous research, challenges remain in extracting semantic features from style-diversified images and optimizing templates within continuous prompt tuning. 
In this study, we employ both NLP and visual prompt tuning to optimize STEM educational content retrieval, enhancing retrieval accuracy and efficiency by adjusting prompt tokens.

\subsection{RAGs and LLMs in Education}
Recent advancements in LLMs such as Chat-GPT~\cite{brown2020language}, Qwen~\cite{qwen2024open}, and LLaMA~\cite{touvron2023llama} have demonstrated strong reasoning and generative capabilities in open-ended question answering, summarization, and tutoring tasks. However, their performance in specialized educational contexts remains limited by several factors, including domain specificity, hallucination, and lack of grounding in accurate curriculum-aligned knowledge.
To mitigate these issues, retrieval-augmented generation (RAG) frameworks~\cite{lewis2020retrieval} have emerged as a promising paradigm. RAG systems typically combine a retriever module that fetches relevant external documents with a generator module that conditions on both the user query and the retrieved evidence to generate answers. Notable variants such as FiD-RAG~\cite{izacard2020leveraging}, Atlas~\cite{izacard2022few}, and REACT-RAG~\cite{yao2023react} demonstrate improved factual accuracy and explainability compared to end-to-end generative models.
Despite this progress, applications of RAG in the educational domain are still in early stages. Existing research mostly focuses on general QA tasks~\cite{welbl2017crowdsourcing} or domain-specific knowledge assistance (e.g., medical~\cite{pal2022medmcqa}, legal~\cite{pipitone2024legalbench}). In education, challenges include the lack of structured multi-modal datasets, complex and hierarchical knowledge representations, and diverse query styles from learners and teachers alike. Moreover, effective RAG systems require alignment between retrieval evidence and generation objectives, which is non-trivial in open-ended learning tasks.

Our work builds upon our previous research~\cite{jia2025uni} by integrating a style-diversified retrieval system (Uni-Retrieval) with a lightweight instruction-tuned LLM, resulting in the Uni-RAG framework.
Unlike existing RAG frameworks, Uni-RAG dynamically augments user queries with domain-specific prompt tokens and retrieves multi-modal educational samples as grounded evidence. This design enables efficient and explainable generation of learning content from diverse educational inputs such as sketches, audio, and low-resolution visuals—providing a scalable and pedagogically aligned solution for intelligent teaching assistance.


\vspace{-0.5em}
\section{Uni-RAG model}
\vspace{-0.35em}

\begin{figure*}[!t]
  \centering
  \includegraphics[width=0.93\linewidth]{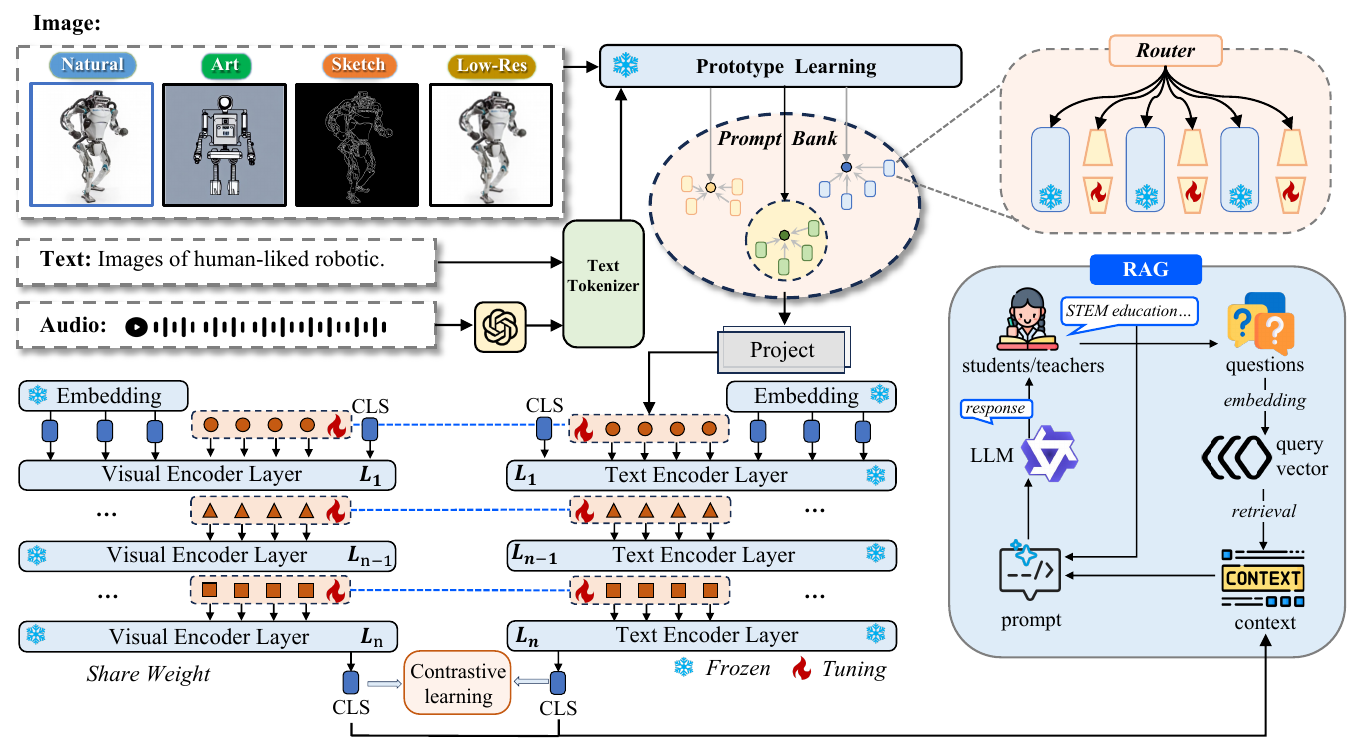}
  \vspace{-3mm}
  \caption{\textbf{The Uni-RAG model's architechture.} Shared prompt tokens are extracted from the Prompt Bank and fed into the input of the feature encoder. Each entry in the Prompt Bank is associated with multiple experts, enabling the representation of diverse style features. After retrieving the top-$k$ relevant items, Uni-RAG concatenates the system prompt with the retrieved content and passes it to the LLM to generate the final explanation for the query.}
  \label{fig:model}
  \vspace{-5mm}
\end{figure*}


In this section, we introduce our Uni-RAG model, as illustrated in Fig. \ref{fig:model}, which consists of five main submodules: a \textbf{prototype learning module} for generating the prototype feature for each content query~(Sec. {\color{red}\ref{subsec:prototype}}), a \textbf{prompt bank with MoE-LoRA adapters module} for saving and adapting style-aware prompts~(Sec. {\color{red}\ref{subsec:prompt_bank}}), a \textbf{feature extractor module} based on different vision-language models for extracting cross-modal features~(Sec.  {\color{red}\ref{subsec:foundation_model}}), a \textbf{RAG module} for domain-aware generation~(Sec. {\color{red}\ref{subsec:rag}}), and a \textbf{training and inference process}~(Sec. {\color{red}\ref{subsec:training_inference}}).

\subsection{Prototype Learning Module} \label{subsec:prototype}
For Uni-RAG, given an input query~(freely based on audio, various styles of images, or text) \( x \subseteq \mathbb{R}^{L*C} \) and an embedding extractor \( f \), map \( x \) into a \( d \)-dimensional shared latent space (Prototype Module) using \( f \). 
For constructing unified multi-modal embedding representations, researchers usually leverage pretrained models with rich semantic priors as modality-specific embedding extractors.
For example, if the input queries focus on style images, we leverage the style embedding extractor~\cite{styler} to map each image into the shared latent space. 
If the query emphasizes the need for more textual information from the context, the text encoder and tokenizer, which are pretrained on large text datasets such as the Pile \cite{pile}, can be utilized. If the query requires alignment with audio information, the audio embedding extractor is be used to transfer the input to the text space.
The input query is embedded as follows:
{\setlength{\abovedisplayskip}{5pt}
\setlength{\belowdisplayskip}{5pt}
\begin{eqnarray}
E_{i} = f(x_{i}), \quad E_{i}\subseteq \mathbb R^{d},i=1,2,\dots,m,
\end{eqnarray}}

\noindent where $E_{i}$ denotes the i-th query. 

\subsection{Prompt Bank with MoE-LoRA Adaptation}
\label{subsec:prompt_bank}
Although static prompt tuning approaches can achieve reasonable performance, they lack flexibility and fail to adapt to subtle variations within a style—particularly in multi-modal or stylistically diverse STEM scenarios. To overcome this limitation, we introduce the Prompt Bank, which stores and dynamically adapts prompt tokens using a Mixture-of-Experts Low-Rank Adapter (\textbf{MoE-LoRA}) mechanism. In contrast to static methods that rely on fixed style embeddings, our design enables the prompt representation to evolve based on the style prototype of the input query.

\paragraph{Prompt Storage and Access} The Prompt Bank maintains a collection of $N$ prompt entries, each consisting of a key-value pair $(k_i, P_i)$ where $k_i$ is a learnable key vector and $P_i$ is a base prompt token in the shared embedding space:
\begin{equation}
\text{Prompt\_Bank} = \{(k_1, P_1), ..., (k_N, P_N)\}.
\end{equation}
Given a query embedding $E_{i}$, the bank retrieves the top-$n$ most similar keys using cosine similarity:
\begin{equation}
K_e = \underset{j\subseteq [1,N]}{\arg\min}\sum_{i=1}^m \gamma(E_{i}, k_{j}).
\end{equation}

\paragraph{MoE-LoRA Prompt Adaptation} To enhance the expressiveness and generalization of the retrieved prompts, we introduce a MoE-LoRA module, which consists of $K$ low-rank expert adapters $\{(A_k, B_k)\}_{k=1}^{K}$ and a lightweight routing network $\phi(\cdot)$.
Each expert operates with low-rank matrices $A^k \in \mathbb{R}^{d \times r}$ and $B^k \in \mathbb{R}^{r \times d}$, where $r$ denotes the rank of the matrix.

Each retrieved prompt $P_i$ is modulated via the router computes expert routing weights $\alpha = \text{softmax}(\phi(P_i)) \in \mathbb{R}^{K}$ and the expert-weighted LoRA adaptation:
\begin{equation}
P_i' = \sum_{k=1}^{K} \alpha_k (P_i^{k} + A_i^{k} B_i^{k}).
\end{equation}
This allows different experts to specialize in encoding distinct domains or modal contexts (e.g., diagrams, handwritten notes, or sketches).

\paragraph{Integration and Composition} The adapted prompts $\{P_{j_1}', ..., P_{j_n}'\}$ are prepended to the input representation as soft prefix tokens:
\begin{equation}
x_p = [\text{CLS}; P_{j_1}'; P_{j_2}'; ...; P_{j_n}'; x_e],
\end{equation}
where $x_e$ denotes the patch/text embeddings of the raw query $x$. These enhanced prompts steer the feature extractor towards more robust and style-aware representations. Moreover, MoE-LoRA module operates with sparse routing and low-rank projection, ensuring parameter and compute efficiency, which makes it particularly well-suited for STEM applications.

\subsection{Feature Extractor}
\label{subsec:foundation_model}
In Uni-RAG, We initialize the vision and text encoders with pretrained weights from OpenCLIP. For tokenization, we use the tokenizer associated with GPT-Neo~\cite{gpt-neo}, which has been trained on the Pile dataset.
What's more, we uses GPT-4o \cite{hurst2024gpt} as the optional choice of the external large language model to convert the audio clips to the text sequence, which facilitates robust and context-aware audio-to-text conversion.

The whole sequence tokens are feed into the feature extractor for training and inference layer by layer. Obtained from the Prompt Bank, visual prompt tokens represent various style information specific to different STEM subjects, while text prompt tokens convey distinct context information about different STEM subjects or different expression about the same subjects. 
The parameters are sharing between visual prompts and text prompts each layer to align vision and text modality. 
In the model training phase, the foundational encoder, tokenizer, and feature extractor are kept frozen to preserve their pretrained semantic space. Only the modules related to the Prompt Bank are updated during training.

\subsection{Retrieval-Augmented Generation}
\label{subsec:rag}
To support downstream educational applications such as explanation generation, context-aware feedback, and multi-modal reasoning, we construct the RAG module. The RAG module fuses retrieved evidence from the Uni-Retrieval~\cite{jia2025uni} pipeline with adapted prompts to form a structured input for an instruction-tuned language model.

\paragraph{Evidence Construction} 
Given a query $x$, we first encode it leveraging the prototype learning module and MoE-LoRA adapted prompt tokens $f_q(x)$. Retrieval is then performed over the embedded educational database $\{f_d(d_i)\}_{i=1}^N$ via cosine similarity function $\mathrm{sim}(\cdot,\cdot)$, and the top-$k$ results are selected as:
\begin{equation}
\mathcal{E}_x = \text{top-k}\left( \mathrm{Sim}\left( f_q(x), \{f_d(d_i)\}_{i=1}^N \right) \right),
\end{equation}
where each retrieved item $d_i$ may consist of textual descriptions, visual exemplars, sketches, or captions. The resulting $\mathcal{E}_x$ serves as semantically aligned evidence for downstream generation.

\paragraph{Generation Input Formation} Let $P^*$ denote the system prompt for LLMs, and let $\mathcal{E}_x$ represent the retrieved results. To effectively guide the language model in generating educationally relevant outputs, we construct the generation input by explicitly structuring the available context as:
\begin{equation}
C_x = [\text{PROMPT}: P^*; \text{EVIDENCE}: \mathcal{E}_x; \text{QUERY}: x].
\end{equation}
This formulation ensures that the model receives: agent-aware guidance from the system prompt $P^*$, semantically aligned support materials from $\mathcal{E}_x$, and the original user query $x$ for grounding.

\paragraph{LLM-Driven Generation} This input $C_x$ is fed into a compact, instruction-tuned large language model such as Qwen3~\cite{yang2025qwen3}, selected for its efficiency and compatibility with multi-lingual and multi-modal scenarios. The language model generates the final response $y$ as:
\begin{equation}
y = \text{LLM}_{\text{Qwen}}(C_x).
\end{equation}
The generation is guided by the prompt, constrained by the retrieved facts, and tailored to the domain context. In practice, the model can generate multiple forms of outputs, such as step-by-step reasoning, style-specific rephrasing, or educational explanations.

\subsection{Training and Inference}
\label{subsec:training_inference}

For the training procedure, in every training step, the visual/context features are extracted from the corresponding encoder $f$ to get the embedding $E_i$.
Then, we compute the similarity between $E_i$ and each key $k_j$, and identify the prompt $P_j$ that best aligns with the embedding.
Besides, the tokenizer and the patch layer map the inputs into sequence $x_t$: 
{\setlength{\abovedisplayskip}{6pt}
\setlength{\belowdisplayskip}{6pt}
\begin{eqnarray}
x_t = \mathrm{Tokenizer/Patch}(x),
\end{eqnarray}}

\noindent where $x_t$ denotes the temp state of features. 
After selecting $n$ prompts following the aforementioned query strategy, the expanded feature embedding $x_p$ is fed into the foundation model $\delta$ and getting the final result $x_f$. We use the $CLS$ token to represent the whole sequence $x_p$ following the settings of classification task:
{\setlength{\abovedisplayskip}{6pt}
\setlength{\belowdisplayskip}{6pt}
\begin{eqnarray}
x_f = \delta(x_p)[:, 0, :].
\end{eqnarray}}

\noindent The triplet loss function $\mathcal{L}$ utilizes the features $x_f$, $x_r$, and $x_h$ of a search query, a retrieval target query, and a negative sample from a different category. 
Minimizing $\mathcal{L}$ brings the correct sample pairs closer together while distancing the negative sample pairs.
With $\mu$ as margin and distance function $d(a,b)=(1-a*b)/(||a||-||b||)$, $\mathcal{L}$ is given as: 
{\setlength{\abovedisplayskip}{6pt}
\setlength{\belowdisplayskip}{6pt}
\begin{multline}
\mathcal L\!=\!max\{0, 
\mu\!+\!d(\delta(x_f),\!\delta(x_r))\!-\!d(\delta(x_f),\!\delta(x_h))\},
\end{multline}}

\noindent where $x_r, x_h$ denotes the embedding of the retrieval object and the negative sample respectively. Moreover, the key in Prompt Bank will be updated with a scale parameter $\lambda$ to weight the second loss function:
{\setlength{\abovedisplayskip}{6pt}
\setlength{\belowdisplayskip}{6pt}
\begin{eqnarray}
\min_{k,p,L} \mathcal L(x_f, x_r, x_h) + \lambda \begin{matrix}\sum_{K_x}\gamma(q(x), k_{si})\end{matrix}.
\end{eqnarray}}

In the generation module of Uni-RAG, the language model is fully frozen to preserve the integrity of the semantic space. During training, Uni-RAG jointly optimizes the Prompt Bank’s key vectors $K$ and prompt tokens $P$, enabling continuous updates and adaptability. Compared to full-parameter fine-tuning, this approach significantly reduces trainable parameters, conserving computational resources and improving training efficiency.

During inference, Uni-RAG first extracts a prototype feature from the unknown-style query input $x$, which is then used to retrieve the most relevant prompt tokens from the Prompt Bank. For unfamiliar styles, the Prompt Bank uses multiple clusters to represent the query jointly. The selected prompt tokens $x_p$ are prepended to the feature extractor $\delta$ for downstream retrieval. To accelerate inference, query embeddings are precomputed and stored in the database.



\section{Experiments}

\begin{table*}[!t]
\centering
\footnotesize
\caption{Retrieval performance comparison on the SER dataset for the STEM education retrieval task. $A \textbf{$\rightarrow$} B$ denotes a retrieval scenario in which a query of type $A$ is used to retrieve an answer of type $B$.}
\renewcommand{\arraystretch}{1.05}  
\setlength{\tabcolsep}{2.75mm}        
{
{
\begin{tabular}{c|c|cc|cc|cc|cc|cc}
    \toprule[1.5pt]
    \multirow{2}{*}{\textbf{\#}} & \multirow{2}{*}{\textbf{Method}} & \multicolumn{2}{c|}{\textbf{Text} \textbf{$\rightarrow$} \textbf{Image}} & \multicolumn{2}{c|}{\textbf{Sketch} \textbf{$\rightarrow$} \textbf{Image}} & \multicolumn{2}{c|}{\textbf{Art} \textbf{$\rightarrow$} \textbf{Image}} & \multicolumn{2}{c|}{\textbf{Low-Res} \textbf{$\rightarrow$} \textbf{Image}} & \multicolumn{2}{c}{\textbf{Audio} \textbf{$\rightarrow$} \textbf{Image}} \\ 
    
    \cmidrule(rl){3-4}\cmidrule(rl){5-6}\cmidrule(rl){7-8}\cmidrule(rl){9-10} \cmidrule(rl){11-12}
    & & {R@1$\uparrow$} & {R@5$\uparrow$} & {R@1$\uparrow$} & {R@5$\uparrow$} & {R@1$\uparrow$} & {R@5$\uparrow$} & {R@1$\uparrow$} & {R@5$\uparrow$} & {R@1$\uparrow$} & {R@5$\uparrow$} \\
    
    \noalign{\hrule height 1.5pt}
    \rowcolor{gray!20}\multicolumn{12}{c}{\it{\textbf{Pretrained Cross-Modality Models}}} \\
    \hline
    1& CLIP             & 54.6 & 78.4 & 47.3 & 68.9 & 46.8 & 71.3 & 53.7 & 72.9 & 30.6 & 52.5 \\
    2& BLIP             & 55.8 & 79.2 & 48.2 & 69.2 & 47.5 & 74.4 & 51.5 & 74.2 & 32.8 & 55.2 \\
    3& BLIP-2           & 63.4 & 89.3 & 74.5 & 89.5 & 52.9 & 84.6 & 72.5 & 90.4 & 39.0 & 64.7 \\
    4& CLIP-Finetune    & 71.4 & 91.4 & 71.0 & 87.0 & 52.2 & 81.6 & 71.2 & 88.1 & 43.2 & 68.7 \\
    5& BLIP-Finetune    & 70.2 & 92.0 & 71.6 & 89.2 & 54.3 & 82.3 & 69.7 & 86.8 & 44.5 & 69.4 \\
    \hline
    \rowcolor{gray!20}\multicolumn{12}{c}{\it{\textbf{Large Multi-Modality Models}}} \\
    \hline
    6& LanguageBind     & 60.2 & 86.9 & 52.8 & 78.4 & 49.0 & 78.4 & 59.1 & 80.2 & 35.3 & 60.9 \\
    7& ImageBind        & 60.8 & 88.3 & 54.2 & 80.5 & 50.6 & 79.8 & 60.2 & 83.7 & 34.2 & 56.4 \\
    8& Unified-IO2      & 67.5 & 89.2 & 59.6 & 84.1 & 55.9 & 82.9 & 64.3 & 84.0 & 37.6 & 63.5 \\
    9& Intern-VL2.5-8B  & 71.8 & 93.2 & 70.6 & 87.1 & 57.3 & 83.5 & 72.0 & 90.4 & 47.5 & 76.7 \\
    \hline
    \rowcolor{gray!20}\multicolumn{12}{c}{\it{\textbf{Style Retrieval Models}}} \\
    \hline
    10& FG-SBIR         & 60.3 & 82.6 & 71.4 & 90.8 & 63.8 & 84.0 & 48.2 & 79.2 & 25.5 & 47.1 \\
    11& SceneTrilogy    & 69.7 & 84.5 & 75.6 & \textbf{96.5} & 71.5 & 92.9 & 68.6 & 85.5 & 26.8 & 49.6 \\
    12& FashionNTM      & 50.4 & 81.3 & 68.9 & 88.6 & 67.1 & 88.9 & 45.6 & 77.5 & 27.4 & 50.5 \\
    \hline
    \rowcolor{gray!20}\multicolumn{12}{c}{\it{\textbf{Cross-Modality Prompt Learning Models}}} \\
    \hline
    13& VPT             & 69.9 & 84.1 & 53.3 & 72.3 & 62.7 & 83.2 & 67.4 & 79.1 & 37.4 & 64.2 \\
    14& CoCoOP          & 72.2 & 86.7 & 53.8 & 74.8 & 66.4 & 87.4 & 70.8 & 81.6 & 39.8 & 67.6 \\
    15& MaPLe           & 73.8 & 87.8 & 62.7 & 78.9 & 67.8 & 89.4 & 71.9 & 86.3 & 42.5 & 70.4 \\
    16& FreestyleRet    & 80.1 & 92.5 & 75.3 & 91.5 & 73.0 & \textbf{98.3} & 78.0 & 90.7 & 46.2 & 73.8 \\
    \hline
    \rowcolor{gray!20}\multicolumn{12}{c}{\it{\textbf{Database-Driven Retrieval Models}}} \\
    \hline
    17& GASKN           & 55.7 & 80.8 & 47.6 & 68.7 & 48.5 & 75.9 & 53.6 & 70.5 & 24.1 & 45.0 \\
    18& SKG             & 57.8 & 82.1 & 45.4 & 65.3 & 49.2 & 76.1 & 56.8 & 75.4 & 24.6 & 46.2 \\
    \noalign{\hrule height 1pt}
    19& Uni-Retrieval & 83.2 & 98.7 & 84.5 & 95.6 & 76.9 & 97.5 & 87.4 & 98.1 & 51.4 & 87.5 \\
    \rowcolor{tkdeblue!60} 20& \textbf{Uni-RAG}(Ours) & \textbf{84.1} & \textbf{99.0} & \textbf{85.1} & 98.1 & \textbf{77.2} & 97.9 & \textbf{89.5} & \textbf{98.7} & \textbf{53.7} & \textbf{89.4} \\
 \bottomrule[1.5pt]
\end{tabular}
}
}
\label{tab:main_results}
\end{table*}

\subsection{Experiments Settings}
\newcommand{\pub}[1]{\color{red}{\tiny{#1}}}
\newcommand{\Frst}[1]{{\textbf{#1}}}
\newcommand{\Scnd}[1]{{\underline{#1}}}

\noindent \textbf{Datasets and Baseline Models:}
We use the SER dataset\cite{jia2025uni} as the primary benchmark and include four additional retrieval datasets to comprehensively evaluate Uni-RAG’s performance. 
Beyond SER, we include four additional datasets: DSR~\cite{li2025freestyleret}, ImageNet-X, SketchCOCO~\cite{gao2020sketchycoco}, and DomainNet~\cite{peng2019moment}. For SketchCOCO and DomainNet, paint/sketch captions are annotated using InternVL-1.5~\cite{Chen_2024_CVPR}. In the prototype learning module, we adopt VGG~\cite{vgg} as the feature extractor. Audio inputs are transcribed using Whisper~\cite{radford2023robust}.
For baseline comparisons, we evaluate two cross-modal models (CLIP~\cite{clip}, BLIP~\cite{blip}), two multi-modal models (LanguageBind~\cite{languagebind}, Unified-IO2~\cite{uio2}), two style retrieval models (SceneTrilogy~\cite{scenetrilogy}, FashionNTM~\cite{fashionntm}), four prompt-learning models (VPT~\cite{jia2022visual}, CoCoOP~\cite{zhou2022conditional}, MaPLe~\cite{khattak2023maple}, FreestyleRet~\cite{li2025freestyleret}), and two database-driven models (GASKN~\cite{GASKN}, MKG~\cite{mkg}).
For the Qwen3 model~\cite{yang2025qwen3}, we select the 0.6B version to maintain algorithmic efficiency.
Cross-modality models are fine-tuned on SER, while prompt-learning models follow VPT settings to adapt to STEM-style diversity. Multi-modal models are evaluated in zero-shot settings for their capacity to handle diverse visual styles. For audio, models with built-in processing use default settings; otherwise, we adopt Uni-RAG’s pipeline.

\noindent \textbf{Evalution Metrics:}
For evalution metric, We evaluate the R@1, R@5 and the inference speed~(ms) on all retrieval datasets. For R@1 and R@5, ``$\uparrow$'' denotes that higher is better. For ms, ``$\downarrow$'' denotes that quicker is better.

\noindent \textbf{Experimental details:}
For the experiments on the SER dataset, Uni-RAG is initialized with OpenCLIP's weights and trained on 8 A100 GPUs with batch size 24 per GPU and 20 training epochs. We use AdamW as the optimizer, set the learning rate to 1e-5 with a linearly warmed up operation in the first epochs and then decayed by the cosine learning rate schedule. The seed is set as 42. What's more, all input images are resized into $224\times224$ resolution and then augmented by normalized operation. All text are padding zero to the max length of 40.

For the fine-tuning CLIP and BLIP models, all experiment settings are the same as Uni-RAG except the learning rate is set as 1e-6. For prompt tuning models, we both use 4 prompt tokens to expand the token sequence. For all transformer-based models, we use the ViT-Large and 24-layers text transformer as the foundation models to keep balance between performance and efficiency.

\begin{table*}[!t]
\centering
\footnotesize
\caption{Retrieval performance for STEM Education Retrieval task. }
\renewcommand{\arraystretch}{1.05}  
\setlength{\tabcolsep}{2.25mm}        
{
\begin{tabular}{c|c|ccccc|cccc|cc|c}
    \toprule[1.5pt]
    
    \textbf{\#} & \textbf{Method} & \textbf{I$\rightarrow$T} & \textbf{S$\rightarrow$T} & \textbf{A$\rightarrow$T} &  \textbf{L$\rightarrow$T} & \textbf{$\mathcal{A}\rightarrow$T} & \textbf{T$\rightarrow$S} & \textbf{T$\rightarrow$A} & \textbf{T$\rightarrow$L} & \textbf{T$\rightarrow \mathcal{A}$} & \textbf{S$\rightarrow$A} & \textbf{S$\rightarrow$L} & \textbf{A$\rightarrow$L} \\
    
    \noalign{\hrule height 1.5pt}
    \rowcolor{gray!20}\multicolumn{14}{c}{\it{\textbf{Metric: R@1$\uparrow$ on SER Dataset}}} \\
    \hline
    1& CLIP             & 47.4 & 38.4 & 37.9 & 38.6 & 27.8 & 38.8 & 37.4 & 35.7 & 28.0 & 36.9 & 34.8 & 31.5 \\
    2& BLIP             & 48.9 & 39.2 & 38.4 & 39.5 & 28.2 & 39.7 & 37.1 & 36.5 & 28.7 & 35.0 & 34.9 & 32.6 \\
    3& BLIP-2           & 69.5 & 67.3 & 66.8 & 68.4 & 31.3 & 66.0 & 65.7 & 66.2 & 36.8 & 59.6 & 58.6 & 53.4 \\
    4& CLIP-Finetune    & 75.7 & 72.4 & 71.3 & 69.8 & 43.7 & 70.2 & 68.5 & 67.7 & 41.4 & 65.4 & 66.8 & 66.3 \\
    5& BLIP-Finetune    & 77.4 & 73.0 & 72.6 & 70.5 & 44.6 & 71.3 & 69.4 & 68.1 & 42.7 & 66.2 & 67.2 & 67.0 \\
    \hline
    6& ImageBind        & 56.2 & 55.4 & 53.5 & 54.0 & 41.9 & 50.2 & 49.3 & 49.8 & 40.1 & 46.6 & 47.5 & 47.1 \\
    7& LanguageBind     & 55.4 & 54.9 & 53.1 & 53.4 & 39.3 & 49.7 & 48.7 & 49.1 & 43.5 & 46.2 & 46.8 & 45.9 \\
    8& Unified-IO2      & 57.3 & 57.2 & 56.3 & 54.5 & 48.0 & 51.1 & 49.9 & 48.6 & 44.8 & 48.0 & 47.2 & 46.8 \\
    9& Intern-VL2.5-8B  & 75.8 & 72.6 & 72.4 & 76.7 & 49.8 & 68.8 & 67.8 & 68.0 & 45.7 & 57.4 & 56.3 & 53.8 \\
    \hline
    10& FG-SBIR         & 68.5 & 65.4 & 63.4 & 64.7 & 28.9 & 68.8 & 68.5 & 65.9 & 26.2 & 64.6 & 65.9 & 63.5 \\
    11& SceneTrilogy    & 72.4 & 76.5 & 70.6 & 71.5 & 29.7 & 69.3 & 69.9 & 68.7 & 27.6 & 65.2 & 66.2 & 64.4 \\
    12& FashionNTM      & 70.6 & 73.3 & 68.9 & 69.6 & 30.2 & 67.1 & 68.0 & 66.5 & 30.5 & 67.5 & 64.8 & 62.4 \\
    \hline
    13& VPT          & 73.9 & 71.8 & 70.4 & 68.7 & 36.5 & 69.0 & 68.2 & 67.4 & 30.5 & 66.6 & 64.5 & 63.8\\
    14& CoCoOP       & 76.5 & 74.7 & 73.4 & 74.0 & 38.2 & 71.4 & 72.3 & 70.8 & 31.0 & 68.9 & 67.2 & 67.3\\
    15& MaPLe      & 78.3 & 75.8 & 75.7 & 74.9 & 45.5 & 72.4 & 69.6 & 69.2 & 46.1 & 68.3 & 67.4 & 65.6 \\
    16& FreestyleRet & 80.8 & 73.5 & \textbf{75.5} & 71.4 & 58.9 & 73.0 & 68.3 & 68.0 & 54.5 & 69.4 & 70.6 & 68.9 \\
    \hline
    17& GASKN        & 53.8 & 52.9 & 52.6 & 50.7 & 24.6 & 49.4 & 47.9 & 46.0 & 24.3 & 47.1 & 47.3 & 45.9\\
    18& SKG          & 54.3 & 51.7 & 50.4 & 51.3 & 25.1 & 48.5 & 46.1 & 45.4 & 26.9 & 46.9 & 47.0 & 45.9\\
    \noalign{\hrule height 1pt}
    19& Uni-Retrieval  & 81.7 & 76.3 & 74.9 & 77.6 & 63.8 & 73.5 & 74.2 & 78.0 & 58.4 & 71.4 & 72.3 & 70.8 \\ 
    \rowcolor{tkdeblue!60} 20& \textbf{Uni-RAG}  & \textbf{82.1} & \textbf{77.0} & \textbf{75.2} & \textbf{77.8} & 65.5 & \textbf{74.1} & \textbf{74.7} & \textbf{78.3} & 60.3 & \textbf{71.6} & \textbf{72.5} & \textbf{71.4}\\ 
 \bottomrule[1.5pt]
\end{tabular}
}
\label{tab:other_results}
\end{table*}

\begin{table}[!t]
    \centering
    \caption{The models inference speed comparison.}
    \vspace{-1mm}
    \resizebox{1.0\columnwidth}!{
    \begin{tabular}{c|cccc}
        \toprule[1.5pt]
        \textbf{Method} & \textbf{Params(M)} & \textbf{Q2I(ms)$\downarrow$} & \textbf{Q2T(ms)$\downarrow$} & \textbf{T$\rightarrow$I(Acc)}$\uparrow$ \\
        \hline
        CLIP & 427M & 68ms & 63ms & 54.6\\
        BLIP & 891M & 62ms & 58ms & 55.8\\
        VPT & 428M & 73ms & 69ms & 69.9\\
        LanguageBind & 1200M & 372ms & 367ms & 60.2\\
        GASKN & 33M & 12ms & 10ms & 55.7\\
        \hline
        Uni-Retrieval & 453M{$_{\textcolor{red}{(+26)}}$} & 77ms{$_{\textcolor{red}{(+9)}}$} & 70ms{$_{\textcolor{red}{(+7)}}$} & 83.2{$_{\textcolor{red}{(+28.6)}}$}\\ 
        \rowcolor{tkdeblue!60} \textbf{Uni-RAG} & 478M{$_{\textcolor{red}{(+51)}}$} & 79ms{$_{\textcolor{red}{(+11)}}$} & 71ms{$_{\textcolor{red}{(+8)}}$} & 84.1{$_{\textcolor{red}{(+29.5)}}$}\\ 
        \bottomrule[1.5pt]
    \end{tabular}}
    \label{tab:speed}
\end{table}

\subsection{Comparison Experiment}

On the SER dataset, Uni-RAG demonstrates superior performance across multiple scenarios with different query styles compared to other baselines, including multi-modality models, cross-modality pre-trained models, and prompt learning models. As shown in Tab.~\ref{tab:main_results}, Tab.~\ref{tab:multi-query} and Tab.~\ref{tab:other_results}, the $T+S\!\rightarrow\!I$ means inputting the text and the style images as the multi-queries and outputting the corresponding images as the target queries, the $\mathcal{A}$ denotes the audio query. The experiment results yield several key observations:

\begin{table}[!t]
    \centering
    \caption{Retrieval performance with multi-style queries simultaneously on SER dataset.}
    \vspace{-1mm}
    \resizebox{1.0\columnwidth}!{
    \begin{tabular}{c|cccc}
        \toprule[1.5pt]
        Method & T$\rightarrow$I & T+S$\rightarrow$I & I$\rightarrow$T & I+S$\rightarrow$T \\
        \hline
        CLIP-Finetune & 54.6 & 55.3{$_{\textcolor{red}{(+0.7)}}$} & 47.4 & 46.6{$_{\textcolor{green}{(-0.8)}}$}\\
        VPT & 69.9 & 72.0{$_{\textcolor{red}{(+2.1)}}$} & 73.9 & 74.1{$_{\textcolor{red}{(+0.2)}}$}\\
        \hline
        Uni-Retrieval & 83.2 & 87.4{$_{\textcolor{red}{(+4.2)}}$} & 81.7 & 83.3{$_{\textcolor{red}{(+1.6)}}$}\\ 
        \rowcolor{tkdeblue!60} \textbf{Uni-RAG} & 84.1 & 88.7{$_{\textcolor{red}{(+4.6)}}$} & 82.1 & 84.5{$_{\textcolor{red}{(+2.4)}}$}\\ 
        \bottomrule[1.5pt]
    \end{tabular}}
    \label{tab:multi-query}
\end{table}

\noindent \textbf{The Uni-RAG achieves the best retrieval performance on the multi-style STEM Education Retrieval task:} This highlights the effectiveness of Uni-RAG in handling complex multi-modal queries. Due to the Prompt Bank's structure, Uni-RAG is a plug-and-play framework that can be highly flexible applied to various multi-modal models and enhance their retrieval capabilities. Line 20 in Tab.~\ref{tab:main_results} provides a substantial performance boost compared to both CLIP and CLIP-Finetune; for example, the R@1 score improves by 12.7\% compared to CLIP-Finetune in the Text-to-Image setting, validating the effectiveness of our framework.

\noindent \textbf{The Prototype module and Prompt Bank significantly outperform full-parameter fine-tuning:} As shown in lines 4-5 and line 20 of Tab.~\ref{tab:main_results}, Uni-RAG surpasses its fine-tuned CLIP counterpart by a large margin. Leveraging the prior knowledge bias introduced by the Prototype module and the efficient memory space of the Prompt Bank, Uni-RAG achieves superior results while tuning less than 10\% of the model’s total parameters. This demonstrates the effectiveness of Uni-RAG’s design in achieving high performance with minimal parameter adjustments.

\noindent \textbf{Uni-RAG can simultaneously perform and mutually enhance traditional text-image retrieval performance:} As shown in Tab.~\ref{tab:multi-query}, when handling text-image retrieval tasks, Uni-RAG allows multi-query inputs as additional references, significantly improving retrieval capability.
For example, compared to the VPT algorithm, Uni-RAG achieves a 14.2\% improvement.
This multi-query input design is not exclusive to Uni-RAG and can also benefit other retrieval models, offering a generalizable approach to enhancing retrieval tasks.

\begin{table*}[!t]
\centering
\footnotesize
\renewcommand{\arraystretch}{1.05}  
\setlength{\tabcolsep}{2.75mm}        
\caption{\textbf{The zero-shot retrieval performance comparison on retrieval datasets.}}
\vspace{-1.5mm}
{
{
\begin{tabular}{l|p{80pt}|cc|cc|cc|cc}
    \toprule[1.5pt]
    \multirow{2}{*}{\textbf{\#}} & \multirow{2}{*}{\textbf{Method}} & \multicolumn{2}{c|}{\textbf{Text} \textbf{$\rightarrow$} \textbf{Image}} & \multicolumn{2}{c|}{\textbf{Sketch} \textbf{$\rightarrow$} \textbf{Image}} & \multicolumn{2}{c|}{\textbf{Art} \textbf{$\rightarrow$} \textbf{Image}} & \multicolumn{2}{c}{\textbf{Low-Res} \textbf{$\rightarrow$} \textbf{Image}} \\ 
    
    \cmidrule(rl){3-4}\cmidrule(rl){5-6}\cmidrule(rl){7-8}\cmidrule(rl){9-10}
    & & {R@1$\uparrow$} & {R@5$\uparrow$} & {R@1$\uparrow$} & {R@5$\uparrow$} & {R@1$\uparrow$} & {R@5$\uparrow$} & {R@1$\uparrow$} & {R@5$\uparrow$} \\

    \noalign{\hrule height 1.5pt}
    \rowcolor{gray!20}\multicolumn{10}{c}{\it{\textbf{Diverse-Style Retrieval Dataset}}} \\
    \hline
    1& ImageBind~\pub{CVPR2023}     & 71.3 & 96.3 & 51.2 & 80.6 & 59.4 & 86.9 & 79.7 & 98.0\\
    2& LanguageBind~\pub{ICLR2024}  & 71.0 & 95.5 & 50.8 & 79.4 & 58.2 & 86.3 & 79.0 & 96.7\\
    3& CoCoOP~\pub{CVPR2022}        & 71.4 & 94.6 & 77.5 & 97.2 & 69.3 & 97.1 & 83.8 & 97.6\\
    4& MaPLe~\pub{CVPR2023}         & 73.1 & 95.9 & 80.3 & 97.9 & 70.6 & 97.2 & 85.9 & 97.7\\
    5& FreestyleRet~\pub{ECCV2024}  & 71.4 & 97.2 & 81.6 & 98.0 & 72.3 & 98.1 & 86.7 & 98.2\\
    \hline
    6& Uni-Retrieval~\pub{ACL2025}  & 82.3 & 97.4 & 82.7 & 98.9 & 75.1 & 98.0 & 91.2 & 98.6 \\
    \rowcolor{tkdeblue!60}
    7& Uni-RAG~(Ours)               & \textbf{84.0} & \textbf{98.2} & \textbf{84.5} & \textbf{99.4} & \textbf{77.6} & \textbf{99.1} & \textbf{91.3} & \textbf{99.4} \\
    \hline
    \rowcolor{gray!20}\multicolumn{10}{c}{\it{\textbf{ImageNet-X Dataset}}} \\
    \hline
    8& GASKN~\pub{ICDE2024}         & 58.7 & 91.8 & 49.2 & 73.3 & 48.9 & 73.6 & 62.5 & 85.1\\
    9& SKG~\pub{ICDE2024}           & 59.2 & 93.0 & 48.5 & 72.4 & 49.1 & 75.4 & 64.3 & 87.2\\
    10& FreestyleRet~\pub{ECCV2024} & 64.8 & 94.3 & 57.7 & 90.5 & 56.4 & 90.2 & 77.4 & 96.5\\
    \hline
    11& Uni-Retrieval~\pub{ACL2025} & 73.9 & 95.7 & 70.6 & 92.5 & 65.3 & 88.4 & 78.6 & 97.0 \\
    \rowcolor{tkdeblue!60}
    12& Uni-RAG~(Ours)              & \textbf{74.4} & \textbf{96.5} & \textbf{72.3} & \textbf{94.6} & \textbf{67.0} & \textbf{91.2} & \textbf{80.5} & \textbf{97.9} \\
    \hline
    \rowcolor{gray!20}\multicolumn{10}{c}{\it{\textbf{DomainNet Dataset}}} \\
    \hline
    13& VPT~\pub{ECCV2022}          & 59.7 & 86.1 & 53.5 & 77.3 & 54.6 & 81.8 & - & -\\
    14& BLIP-Finetune               & 65.3 & 94.2 & 71.4 & 89.7 & 54.3 & 82.3 & - & -\\
    15& FreestyleRet~\pub{ECCV2024} & 70.2 & 95.2 & 75.2 & 93.2 & 73.1 & 92.6 & - & -\\
    \hline
    16& Uni-Retrieval~\pub{ACL2025} & 70.7 & 96.0 & 77.6 & 94.1 & 73.4 & 92.9 & - & - \\
    \rowcolor{tkdeblue!60}
    17& Uni-RAG~(Ours)              & \textbf{71.2} & \textbf{96.4} & \textbf{79.3} & \textbf{96.5} & \textbf{75.0} & \textbf{94.5} & - & - \\
    \hline
    \rowcolor{gray!20}\multicolumn{10}{c}{\it{\textbf{SketchCOCO Dataset}}} \\
    \hline
    18& MaPLe~\pub{CVPR2023}        & 26.4 & 53.9 & 18.0 & 48.3 & - & - & - & -\\
    19& FG-SBIR~\pub{ECCV2022}      & 27.3 & 56.5 & 19.6 & 49.1 & - & - & - & -\\
    20& SceneTrilogy~\pub{CVPR2023} & 30.6 & 65.8 & 22.5 & 51.6 & - & - & - & -\\
    21& FreestyleRet~\pub{ECCV2024} & 31.5 & 67.3 & 29.6 & 56.1 & - & - & - & -\\
    \hline
    22& Uni-Retrieval~\pub{ACL2025} & 34.7 & 71.6 & 30.2 & 60.4 & - & - & - & - \\
    \rowcolor{tkdeblue!60}
    23& Uni-RAG~(Ours)              & \textbf{35.0} & \textbf{72.2} & \textbf{30.6} & \textbf{61.3} & - & - & - & - \\
 \bottomrule[1.5pt]
\end{tabular}
}
}
\label{tab:multi_results}
\end{table*}

\begin{table}[!t]
    \centering
    \caption{Abalation study on prompt bank and feature extractor.}
    \resizebox{0.6\columnwidth}!{
    \begin{tabular}{c|cccc}
        \toprule[1pt]
        \multirow{2}{*}{Method} & \multicolumn{4}{c}{Prompt Bank Size} \\
        \cmidrule(rl){2-5}
         & 4 & 8 & 16 & 32 \\
        \hline
        ResNet & 68.8 & 73.2 & 79.5 & 78.5 \\
        VGG & 77.5 & 80.6 & \textbf{84.1} & 83.0 \\
        \bottomrule[1pt]
    \end{tabular}}
    \vspace{-1.5em}
    \label{tab:prompt-bank}
\end{table}

\noindent \textbf{Uni-RAG still requires a unified audio encoder to effectively align audio features with the visual-text semantic space:} Among all retrieval modalities, audio-based performance remains the weakest across models. This limitation is largely due to variations in speaker accent, speaking rate, and content complexity. 
As shown in the last column of Tab.~\ref{tab:main_results}, a substantial performance gap persists in audio-based retrieval. Despite this, Uni-RAG achieves the highest overall accuracy on the SER dataset, with an R@5 score of 89.4\%, outperforming all baselines. This result highlights the robustness of our cross-modal alignment strategy, which enables Uni-RAG to maintain strong retrieval performance even under challenging audio conditions.

In addition to accuracy, inference speed is a crucial metric for evaluating retrieval models. As shown in Tab.~\ref{tab:speed}, Uni-RAG adds just 11ms per search iteration. 
Compared to GASKN, Uni-RAG demonstrates significantly stronger retrieval performance than traditional database-driven methods. 
Additionally, when compared to other cross-modality methods, Uni-RAG excels in both tuning efficiency and retrieval accuracy, further validating its effectiveness and scalability.

\subsection{Ablation Study}

To quantitatively assess the role of prompts in Uni-RAG, we conducted ablation studies examining prompt insertion type, prompt token count, and Prompt Bank size across five style metrics. These experiments aimed to evaluate the impact of prompt configurations on style-diversified STEM education retrieval, providing insights into performance and adaptability. Following VPT's design~\cite{jia2022visual}, in the shallow setting, prompt tokens are inserted only at the first layer, whereas the deep setting inserts them across all layers. The token number indicates the repetition count of each prompt token, and the Prompt Bank size refers to the total number of query–key pairs.

\begin{table}[!t]
    \centering
    \caption{Ablation study for prompt settings.}
    \resizebox{1.0\columnwidth}!{
    \begin{tabular}{c|c|c|ccccc}
        \toprule[1.5pt]
        \textbf{\#} & \textbf{Type} & \textbf{Token-Num} & \textbf{T\textbf{$\rightarrow$}I} & \textbf{S\textbf{$\rightarrow$}I} & \textbf{A\textbf{$\rightarrow$}I} & \textbf{L\textbf{$\rightarrow$}I} & \textbf{$\mathcal{A}\rightarrow$I}\\ 
        \hline
        1 & Deep & 1 & 72.0 & 78.3 & 73.2 & 80.6 & 43.8 \\
        2 & Deep & 2 & 77.1 & 81.2 & 75.5 & 85.8 & 49.6 \\
        3 & Deep & 8 & \textbf{83.3} & 82.7 & 76.5 & 87.0 & 53.5 \\
        4 & Shallow & 4 & 68.2 & 75.6 & 70.4 & 77.3 & 46.9 \\
        \hline
        \rowcolor{tkdeblue!60} 
        5 & Deep & 4 & 83.2 & \textbf{84.5} & \textbf{76.9} & \textbf{87.4} & \textbf{53.7} \\
        \bottomrule[1.5pt]
    \end{tabular}
    }
    \label{ablation:prompt}
    \vspace{-5mm}
\end{table}

As shown in lines 4–5 of Tab.~\ref{ablation:prompt}, the deep insertion type consistently outperforms the shallow one, with an average performance gain of 9.46\%. Lines 1–3 and 5 further illustrate that increasing the prompt token count beyond four repetitions yields minimal performance improvement while significantly increasing the number of tunable parameters. This leads to slower tuning and inference. Therefore, inserting four prompt tokens at each layer achieves a balanced trade-off between accuracy and efficiency, and was adopted as Uni-RAG's default configuration.

A similar trend is observed with the Prompt Bank size. As shown in Tab.~\ref{tab:prompt-bank}, doubling the number of query–key pairs does not yield additional performance gains. The best results are achieved when the Prompt Bank size is set to 16, reinforcing the importance of carefully balancing model complexity and computational cost.

We also evaluated Uni-RAG’s zero-shot retrieval performance on several other multi-style datasets. As shown in Tab.~\ref{tab:multi_results}, we compared Uni-RAG against various baseline models across three datasets: the DSR, DomainNet, and SketchCOCO dataset, each representing distinct domains of style-based queries. 
As shown in Tab.~\ref{tab:multi_results}, Uni-RAG exhibits strong zero-shot performance across multiple diverse datasets. For instance, it outperforms the FreestyleRet model by 12.6\% in R@1 on the DSR dataset, underscoring its effectiveness in retrieving relevant information from previously unseen sources.
Furthermore, Uni-RAG consistently surpasses all baseline models—including large multi-modality models, style retrieval models, prompt learning approaches, and database-driven frameworks—across datasets with varying style representations, demonstrating its broad generalization capability and style-adaptive retrieval performance.
This performance underscores Uni-RAG’s scalability and robustness, significantly enhancing the adaptability and effectiveness of existing retrieval models in handling diverse and unstructured data domains.

\subsection{Visualization Result}

\begin{figure*}[!t]
    \centering
    \includegraphics[width=0.99\linewidth]{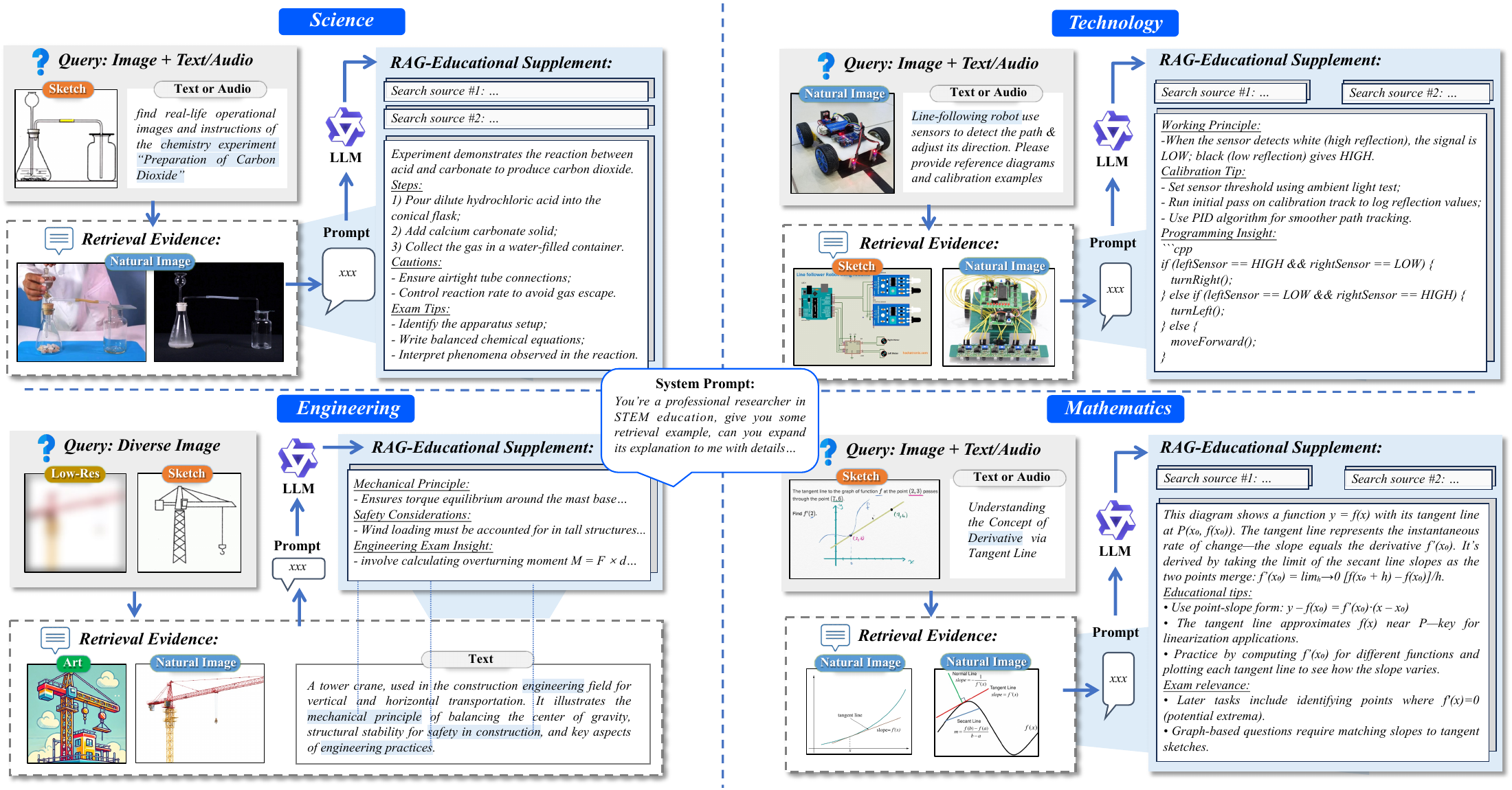}
   \vspace{-0.95mm}
    \caption{The case study for our Uni-RAG with the FreestyleRet baseline involves examples from diverse disciplines within Science, Technology, Engineering, and Mathematics.}
    \label{fig:visualization}
\end{figure*}

In Fig.~\ref{fig:visualization}, we present visualizations of style-diverse queries and their corresponding retrieval results generated by Uni-RAG. Each quadrant corresponds to one of the four STEM domains—Science, Technology, Engineering, and Mathematics. The input modalities span natural images, low-resolution images, hand-drawn sketches, and audio/text descriptions, reflecting the varied formats encountered in real-world educational settings.

Uni-RAG effectively interprets these heterogeneous inputs and pedagogical intents, retrieving both naturalistic and schematic representations from a knowledge base or external corpus. For instance, in the engineering domain, a query regarding a crane retrieves both realistic and artistic depictions, enriching conceptual understanding. In cross-modal scenarios, such as chemistry or robotics, the system successfully aligns sketches with high-quality images—critical for resource-constrained settings like whiteboard drawings or handwritten inputs.

Following retrieval, Uni-RAG prompts a LLM using a unified system prompt designed to emulate a professional STEM educator. This prompt elicits structured, curriculum-aligned explanations tailored to each domain. For science, Uni-RAG provides step-by-step explanations of chemical reactions, safety precautions, and exam preparation strategies. For technology, it illustrates working principles, calibration procedures, and annotated source code. In engineering, Uni-RAG explains mechanical concepts and structural safety considerations, such as wind load analysis. For mathematics, it offers interpretations of derivatives, graph-reading strategies, and common exam question formats. This integration of precise retrieval and generative explanation underscores Uni-RAG’s potential for educational support across diverse STEM contexts.
\section{Conclusion}

To unify multi-style retrieval and generation in educational settings, we proposed Uni-RAG, an enhanced framework built upon Uni-Retrieval. Uni-RAG integrates a lightweight, plug-and-play prompt bank powered by MoE-LoRA, enabling efficient and adaptive prompting for various educational tasks. It also incorporates a compact yet powerful language model, designed to balance performance and computational efficiency. Compared to existing retrieval-based and RAG models, Uni-RAG achieves superior performance while requiring fewer than 50M activated parameters, highlighting its effectiveness in low-resource and real-time educational applications.
Furthermore, the training and deployment costs of Uni-RAG are substantially lower than those of existing large retrieval models, making it a more economical and practical solution for educational scenarios.
We hope that Uni-RAG can inspire new possibilities within the research community by offering an effective and accessible approach to retrieval in STEM education and beyond. 

Looking forward, we plan to extend Uni-RAG to the video modality, enabling it to handle common educational tasks such as video summarization, captioning, and retrieval, further enhancing its applicability across a wide range of multi-modal educational scenarios.

{
    \bibliographystyle{IEEEtran}
    \bibliography{reference}

\begin{thebibliography}{10}
\providecommand{\url}[1]{#1}
\csname url@samestyle\endcsname
\providecommand{\newblock}{\relax}
\providecommand{\bibinfo}[2]{#2}
\providecommand{\BIBentrySTDinterwordspacing}{\spaceskip=0pt\relax}
\providecommand{\BIBentryALTinterwordstretchfactor}{4}
\providecommand{\BIBentryALTinterwordspacing}{\spaceskip=\fontdimen2\font plus
\BIBentryALTinterwordstretchfactor\fontdimen3\font minus \fontdimen4\font\relax}
\providecommand{\BIBforeignlanguage}[2]{{%
\expandafter\ifx\csname l@#1\endcsname\relax
\typeout{** WARNING: IEEEtran.bst: No hyphenation pattern has been}%
\typeout{** loaded for the language `#1'. Using the pattern for}%
\typeout{** the default language instead.}%
\else
\language=\csname l@#1\endcsname
\fi
#2}}
\providecommand{\BIBdecl}{\relax}
\BIBdecl

\bibitem{hwang2020vision}
G.-J. Hwang, H.~Xie, B.~W. Wah, and D.~Ga{\v{s}}evi{\'c}, ``Vision, challenges, roles and research issues of artificial intelligence in education,'' p. 100001, 2020.

\bibitem{wang2023improving}
J.~Wang, X.~Liang, B.~Whitney, J.~Chen, Q.~Gong, X.~He \emph{et~al.}, ``Improving progressive retrieval for hpc scientific data using deep neural network,'' in \emph{2023 IEEE 39th International Conference on Data Engineering}, 2023, pp. 2727--2739.

\bibitem{yang2025qwen3}
A.~Yang, A.~Li, B.~Yang, B.~Zhang, B.~Hui, B.~Zheng, B.~Yu, C.~Gao, C.~Huang, C.~Lv \emph{et~al.}, ``Qwen3 technical report,'' \emph{arXiv preprint arXiv:2505.09388}, 2025.

\bibitem{prompt14}
X.~Liu, Y.~Zheng, Z.~Du, M.~Ding, Y.~Qian, Z.~Yang, and J.~Tang, ``Gpt understands, too,'' \emph{AI Open}, 2023.

\bibitem{intro3}
T.~Shen, X.~Geng, C.~Tao, C.~Xu, G.~Long, K.~Zhang, and D.~Jiang, ``Unifier: A unified retriever for large-scale retrieval,'' in \emph{Proceedings of the 29th ACM SIGKDD Conference on Knowledge Discovery and Data Mining}, 2023, pp. 4787--4799.

\bibitem{li2024alleviating}
T.~Li, X.~Yang, Y.~Ke, B.~Wang, Y.~Liu, and J.~Xu, ``Alleviating the inconsistency of multimodal data in cross-modal retrieval,'' in \emph{2024 IEEE 40th International Conference on Data Engineering}, 2024, pp. 4643--4656.

\bibitem{intro12}
M.~Williams-Lekuona, G.~Cosma, and I.~Phillips, ``A framework for enabling unpaired multi-modal learning for deep cross-modal hashing retrieval,'' \emph{Journal of Imaging}, vol.~8, no.~12, p. 328, 2022.

\bibitem{intro11}
H.~Zhou, Y.~Hu, S.~Liu, G.~Zhou, J.~Xu, A.~Chen, Y.~Wang, L.~Li, and Y.~Hu, ``A precise framework for rice leaf disease image--text retrieval using fhtw-net,'' \emph{Plant Phenomics}, vol.~6, p. 0168, 2024.

\bibitem{li2025freestyleret}
H.~Li, Y.~Jia, P.~Jin, Z.~Cheng, K.~Li, J.~Sui, C.~Liu, and L.~Yuan, ``Freestyleret: Retrieving images from style-diversified queries,'' in \emph{European Conference on Computer Vision}, 2025, pp. 258--274.

\bibitem{intro6}
Q.~Yang, M.~Ye, Z.~Cai, K.~Su, and B.~Du, ``Composed image retrieval via cross relation network with hierarchical aggregation transformer,'' \emph{IEEE Transactions on Image Processing}, 2023.

\bibitem{jia2025uni}
Y.~Jia, X.~Wu, H.~Li, Q.~Zhang, Y.~Hu, S.~Zhao, and W.~Fan, ``Uni-retrieval: A multi-style retrieval framework for stem's education,'' in \emph{Proceedings of the 63rd Annual Meeting of the Association for Computational Linguistics}, 2025.

\bibitem{9392366}
Y.~Zhang and Q.~Yang, ``A survey on multi-task learning,'' \emph{IEEE Transactions on Knowledge and Data Engineering}, vol.~34, no.~12, pp. 5586--5609, 2022.

\bibitem{xiao2025exploring}
L.~Xiao, R.~Mao, S.~Zhao, Q.~Lin, Y.~Jia, L.~He, and E.~Cambria, ``Exploring cognitive and aesthetic causality for multimodal aspect-based sentiment analysis,'' \emph{IEEE Transactions on Affective Computing}, 2025.

\bibitem{jia2025seeing}
Y.~Jia, J.~Xie, S.~Jivaganesh, H.~Li, X.~Wu, and M.~Zhang, ``Seeing sound, hearing sight: Uncovering modality bias and conflict of ai models in sound localization,'' \emph{arXiv preprint arXiv:2505.11217}, 2025.

\bibitem{jia2025robust}
Y.~Jia, X.~Wu, Q.~Zhang, Y.~Qin, L.~Xiao, and S.~Zhao, ``Towards robust evaluation of stem education: Leveraging mllms in project-based learning,'' \emph{ResearchGate}, 2025.

\bibitem{farahani2021brief}
A.~Farahani, S.~Voghoei, K.~Rasheed, and H.~R. Arabnia, ``A brief review of domain adaptation,'' \emph{Advances in data science and information engineering: proceedings from ICDATA 2020 and IKE 2020}, pp. 877--894, 2021.

\bibitem{zhou2022domain}
K.~Zhou, Z.~Liu, Y.~Qiao, T.~Xiang, and C.~C. Loy, ``Domain generalization: A survey,'' \emph{IEEE Transactions on Pattern Analysis and Machine Intelligence}, vol.~45, no.~4, pp. 4396--4415, 2022.

\bibitem{ruder2017overview}
S.~Ruder, ``An overview of multi-task learning in deep neural networks,'' \emph{arXiv preprint arXiv:1706.05098}, 2017.

\bibitem{kokkinos2017ubernet}
I.~Kokkinos, ``Ubernet: Training a universal convolutional neural network for low-, mid-, and high-level vision using diverse datasets and limited memory,'' in \emph{Proceedings of the IEEE conference on computer vision and pattern recognition}, 2017, pp. 6129--6138.

\bibitem{yang2016trace}
Y.~Yang and T.~M. Hospedales, ``Trace norm regularised deep multi-task learning,'' \emph{arXiv preprint arXiv:1606.04038}, 2016.

\bibitem{zhou2022conditional}
K.~Zhou, J.~Yang, C.~C. Loy, and Z.~Liu, ``Conditional prompt learning for vision-language models,'' in \emph{Proceedings of the IEEE/CVF conference on computer vision and pattern recognition}, 2022, pp. 16\,816--16\,825.

\bibitem{khattak2023maple}
M.~U. Khattak, H.~Rasheed, M.~Maaz, S.~Khan \emph{et~al.}, ``Maple: Multi-modal prompt learning,'' in \emph{Proceedings of the IEEE/CVF Conference on Computer Vision and Pattern Recognition}, 2023, pp. 19\,113--19\,122.

\bibitem{chen2022deep}
W.~Chen, Y.~Liu, W.~Wang, E.~M. Bakker, T.~Georgiou, P.~Fieguth, L.~Liu, and M.~S. Lew, ``Deep learning for instance retrieval: A survey,'' \emph{IEEE Transactions on Pattern Analysis and Machine Intelligence}, vol.~45, no.~6, pp. 7270--7292, 2022.

\bibitem{li2011text}
W.~Li, L.~Duan, D.~Xu, and I.~W.-H. Tsang, ``Text-based image retrieval using progressive multi-instance learning,'' in \emph{2011 international conference on computer vision}.\hskip 1em plus 0.5em minus 0.4em\relax IEEE, 2011, pp. 2049--2055.

\bibitem{neculai2022probabilistic}
A.~Neculai, Y.~Chen, and Z.~Akata, ``Probabilistic compositional embeddings for multimodal image retrieval,'' in \emph{Proceedings of the IEEE/CVF conference on computer vision and pattern recognition}, 2022, pp. 4547--4557.

\bibitem{lee-etal-2024-interactive}
S.~Lee, S.~Yu, J.~Park, J.~Yi, and S.~Yoon, ``Interactive text-to-image retrieval with large language models: A plug-and-play approach,'' in \emph{Proceedings of the 62nd Annual Meeting of the Association for Computational Linguistics}, L.-W. Ku, A.~Martins, and V.~Srikumar, Eds., 2024, pp. 791--809.

\bibitem{chowdhury2022fs}
P.~N. Chowdhury, A.~Sain, A.~K. Bhunia, T.~Xiang, Y.~Gryaditskaya, and Y.-Z. Song, ``Fs-coco: Towards understanding of freehand sketches of common objects in context,'' in \emph{European conference on computer vision}.\hskip 1em plus 0.5em minus 0.4em\relax Springer, 2022, pp. 253--270.

\bibitem{chowdhury2023scenetrilogy}
P.~N. Chowdhury, A.~K. Bhunia, A.~Sain, S.~Koley, T.~Xiang, and Y.-Z. Song, ``Scenetrilogy: On human scene-sketch and its complementarity with photo and text,'' in \emph{Proceedings of the IEEE/CVF Conference on Computer Vision and Pattern Recognition}, 2023, pp. 10\,972--10\,983.

\bibitem{johnson2015image}
J.~Johnson, R.~Krishna, M.~Stark, L.-J. Li, D.~Shamma, M.~Bernstein, and L.~Fei-Fei, ``Image retrieval using scene graphs,'' in \emph{Proceedings of the IEEE conference on computer vision and pattern recognition}, 2015, pp. 3668--3678.

\bibitem{li-etal-2024-generative}
Y.~Li, W.~Wang, L.~Qu, L.~Nie, W.~Li, and T.-S. Chua, ``Generative cross-modal retrieval: Memorizing images in multimodal language models for retrieval and beyond,'' in \emph{Proceedings of the 62nd Annual Meeting of the Association for Computational Linguistics}, 2024, pp. 11\,851--11\,861.

\bibitem{zhou-etal-2024-vista}
J.~Zhou, Z.~Liu, S.~Xiao, B.~Zhao, and Y.~Xiong, ``{VISTA}: Visualized text embedding for universal multi-modal retrieval,'' in \emph{Proceedings of the 62nd Annual Meeting of the Association for Computational Linguistics}, 2024, pp. 3185--3200.

\bibitem{brown2020language}
T.~Brown, B.~Mann, N.~Ryder, M.~Subbiah, J.~Kaplan, P.~Dhariwal, A.~Neelakantan, P.~Shyam, G.~Sastry, A.~Askell \emph{et~al.}, ``Language models are few-shot learners,'' in \emph{Proceedings of the 34th International Conference on Neural Information Processing Systems}, 2020.

\bibitem{li2023weakly}
H.~Li, J.~Huang, P.~Jin, G.~Song, Q.~Wu, and J.~Chen, ``Weakly-supervised 3d spatial reasoning for text-based visual question answering,'' in \emph{Transactions on Image Processing}.\hskip 1em plus 0.5em minus 0.4em\relax IEEE, 2023, pp. 3367--3382.

\bibitem{zhao2023prompt}
S.~Zhao, J.~Wen, A.~Luu, J.~Zhao, and J.~Fu, ``Prompt as triggers for backdoor attack: Examining the vulnerability in language models,'' in \emph{Proceedings of the 2023 Conference on Empirical Methods in Natural Language Processing}, 2023, pp. 12\,303--12\,317.

\bibitem{zhao2024unlearning}
S.~Zhao, X.~Wu, C.-D. Nguyen, M.~Jia, Y.~Feng, and L.~A. Tuan, ``Unlearning backdoor attacks for llms with weak-to-strong knowledge distillation,'' in \emph{Proceedings of the 63rd Annual Meeting of the Association for Computational Linguistics}, 2025.

\bibitem{prompt3}
B.~Lester, R.~Al-Rfou, and N.~Constant, ``The power of scale for parameter-efficient prompt tuning,'' \emph{arXiv preprint arXiv:2104.08691}, 2021.

\bibitem{dong2024survey}
Q.~Dong, L.~Li, D.~Dai, C.~Zheng, J.~Ma, R.~Li, H.~Xia, J.~Xu, Z.~Wu, B.~Chang \emph{et~al.}, ``A survey on in-context learning,'' in \emph{Proceedings of the 2024 Conference on Empirical Methods in Natural Language Processing}, 2024, pp. 1107--1128.

\bibitem{zhao2024universal}
S.~Zhao, M.~Jia, L.~A. Tuan, F.~Pan, and J.~Wen, ``Universal vulnerabilities in large language models: Backdoor attacks for in-context learning,'' in \emph{Proceedings of the 2024 Conference on Empirical Methods in Natural Language Processing}, 2024, pp. 11\,507--11\,522.

\bibitem{wang2024pandalm}
Y.~Wang, Z.~Yu, W.~Yao, Z.~Zeng, L.~Yang, C.~Wang \emph{et~al.}, ``Pandalm: An automatic evaluation benchmark for llm instruction tuning optimization,'' in \emph{The Twelfth International Conference on Learning Representations}, 2024.

\bibitem{prompt4}
J.~Wei, X.~Wang, D.~Schuurmans, M.~Bosma, F.~Xia, E.~Chi, Q.~V. Le, D.~Zhou \emph{et~al.}, ``Chain-of-thought prompting elicits reasoning in large language models,'' \emph{Advances in neural information processing systems}, vol.~35, pp. 24\,824--24\,837, 2022.

\bibitem{lester2021power}
B.~Lester, R.~Al-Rfou, and N.~Constant, ``The power of scale for parameter-efficient prompt tuning,'' in \emph{Proceedings of the 2021 Conference on Empirical Methods in Natural Language Processing}, 2021, pp. 3045--3059.

\bibitem{long2024prompt}
D.~Long, Y.~Zhao, H.~Brown, Y.~Xie, J.~Zhao, N.~Chen, K.~Kawaguchi, M.~Shieh, and J.~He, ``Prompt optimization via adversarial in-context learning,'' in \emph{Proceedings of the 62nd Annual Meeting of the Association for Computational Linguistics}, 2024, pp. 7308--7327.

\bibitem{shen2024multitask}
S.~Shen, S.~Yang, T.~Zhang, B.~Zhai, J.~E. Gonzalez, K.~Keutzer, and T.~Darrell, ``Multitask vision-language prompt tuning,'' in \emph{Proceedings of the IEEE/CVF Winter Conference on Applications of Computer Vision}, 2024, pp. 5656--5667.

\bibitem{jia2022visual}
M.~Jia, L.~Tang, B.-C. Chen, C.~Cardie, S.~Belongie \emph{et~al.}, ``Visual prompt tuning,'' in \emph{European Conference on Computer Vision}, 2022, pp. 709--727.

\bibitem{wang2022learning}
Z.~Wang, Z.~Zhang, C.-Y. Lee, H.~Zhang, R.~Sun, X.~Ren, G.~Su, V.~Perot, J.~Dy, and T.~Pfister, ``Learning to prompt for continual learning,'' in \emph{Proceedings of the IEEE/CVF conference on computer vision and pattern recognition}, 2022, pp. 139--149.

\bibitem{ge2023domain}
C.~Ge, R.~Huang, M.~Xie, Z.~Lai, S.~Song, S.~Li, and G.~Huang, ``Domain adaptation via prompt learning,'' \emph{IEEE Transactions on Neural Networks and Learning Systems}, 2023.

\bibitem{nie2023pro}
X.~Nie, B.~Ni, J.~Chang, G.~Meng, C.~Huo \emph{et~al.}, ``Pro-tuning: Unified prompt tuning for vision tasks,'' \emph{IEEE Transactions on Circuits and Systems for Video Technology}, 2023.

\bibitem{cho2023distribution}
E.~Cho, J.~Kim, and H.~J. Kim, ``Distribution-aware prompt tuning for vision-language models,'' in \emph{Proceedings of the IEEE/CVF International Conference on Computer Vision}, 2023, pp. 22\,004--22\,013.

\bibitem{qwen2024open}
J.~Bai, S.~Bai, Y.~Chu, Z.~Cui, K.~Dang, X.~Deng, Y.~Fan, W.~Ge, Y.~Han, F.~Huang \emph{et~al.}, ``Qwen technical report,'' \emph{arXiv preprint arXiv:2309.16609}, 2023.

\bibitem{touvron2023llama}
H.~Touvron, T.~Lavril, G.~Izacard, X.~Martinet, M.-A. Lachaux \emph{et~al.}, ``Llama: Open and efficient foundation language models,'' \emph{arXiv preprint arXiv:2302.13971}, 2023.

\bibitem{lewis2020retrieval}
P.~Lewis, E.~Perez, A.~Piktus, F.~Petroni, V.~Karpukhin, N.~Goyal, H.~K{\"u}ttler, M.~Lewis, W.-t. Yih, T.~Rockt{\"a}schel \emph{et~al.}, ``Retrieval-augmented generation for knowledge-intensive nlp tasks,'' \emph{Advances in neural information processing systems}, vol.~33, pp. 9459--9474, 2020.

\bibitem{izacard2020leveraging}
G.~Izacard and E.~Grave, ``Leveraging passage retrieval with generative models for open domain question answering,'' \emph{arXiv preprint arXiv:2007.01282}, 2020.

\bibitem{izacard2022few}
G.~Izacard, P.~Lewis, M.~Lomeli, L.~Hosseini, F.~Petroni, T.~Schick, J.~Dwivedi-Yu, A.~Joulin, S.~Riedel, and E.~Grave, ``Few-shot learning with retrieval augmented language models,'' \emph{arXiv preprint arXiv:2208.03299}, vol.~1, no.~2, p.~4, 2022.

\bibitem{yao2023react}
S.~Yao, J.~Zhao, D.~Yu, N.~Du, I.~Shafran, K.~Narasimhan, and Y.~Cao, ``React: Synergizing reasoning and acting in language models,'' in \emph{International Conference on Learning Representations (ICLR)}, 2023.

\bibitem{welbl2017crowdsourcing}
J.~Welbl, N.~F. Liu, and M.~Gardner, ``Crowdsourcing multiple choice science questions,'' \emph{arXiv preprint arXiv:1707.06209}, 2017.

\bibitem{pal2022medmcqa}
A.~Pal, L.~K. Umapathi, and M.~Sankarasubbu, ``Medmcqa: A large-scale multi-subject multi-choice dataset for medical domain question answering,'' in \emph{Conference on health, inference, and learning}.\hskip 1em plus 0.5em minus 0.4em\relax PMLR, 2022, pp. 248--260.

\bibitem{pipitone2024legalbench}
N.~Pipitone and G.~H. Alami, ``Legalbench-rag: A benchmark for retrieval-augmented generation in the legal domain,'' \emph{arXiv preprint arXiv:2408.10343}, 2024.

\bibitem{styler}
Y.~Tao, ``Image style transfer based on vgg neural network model,'' in \emph{2022 IEEE International Conference on Advances in Electrical Engineering and Computer Applications (AEECA)}, 2022, pp. 1475--1482.

\bibitem{pile}
L.~Gao, S.~Biderman, S.~Black, L.~Golding, T.~Hoppe, C.~Foster \emph{et~al.}, ``The {P}ile: An 800gb dataset of diverse text for language modeling,'' \emph{arXiv preprint arXiv:2101.00027}, 2020.

\bibitem{gpt-neo}
S.~Black, G.~Leo, P.~Wang, C.~Leahy, and S.~Biderman, ``{GPT-Neo: Large Scale Autoregressive Language Modeling with Mesh-Tensorflow},'' Mar. 2021.

\bibitem{hurst2024gpt}
A.~Hurst, A.~Lerer, A.~P. Goucher, A.~Perelman, A.~Ramesh, A.~Clark \emph{et~al.}, ``Gpt-4o system card,'' \emph{arXiv preprint arXiv:2410.21276}, 2024.

\bibitem{gao2020sketchycoco}
C.~Gao, Q.~Liu, Q.~Xu, L.~Wang, J.~Liu, and C.~Zou, ``Sketchycoco: Image generation from freehand scene sketches,'' in \emph{Proceedings of the IEEE/CVF conference on computer vision and pattern recognition}, 2020, pp. 5174--5183.

\bibitem{peng2019moment}
X.~Peng, Q.~Bai, X.~Xia, Z.~Huang, K.~Saenko, and B.~Wang, ``Moment matching for multi-source domain adaptation,'' in \emph{Proceedings of the IEEE/CVF international conference on computer vision}, 2019, pp. 1406--1415.

\bibitem{Chen_2024_CVPR}
Z.~Chen, J.~Wu, W.~Wang, W.~Su, G.~Chen, S.~Xing, M.~Zhong \emph{et~al.}, ``Internvl: Scaling up vision foundation models and aligning for generic visual-linguistic tasks,'' in \emph{Proceedings of the IEEE/CVF Conference on Computer Vision and Pattern Recognition (CVPR)}, June 2024, pp. 24\,185--24\,198.

\bibitem{vgg}
K.~Simonyan, ``Very deep convolutional networks for large-scale image recognition,'' \emph{arXiv preprint arXiv:1409.1556}, 2014.

\bibitem{radford2023robust}
A.~Radford, J.~W. Kim, T.~Xu, G.~Brockman, C.~McLeavey, and I.~Sutskever, ``Robust speech recognition via large-scale weak supervision,'' in \emph{International conference on machine learning}.\hskip 1em plus 0.5em minus 0.4em\relax PMLR, 2023, pp. 28\,492--28\,518.

\bibitem{clip}
A.~Radford, J.~W. Kim, C.~Hallacy, A.~Ramesh \emph{et~al.}, ``Learning transferable visual models from natural language supervision,'' in \emph{International conference on machine learning}, 2021, pp. 8748--8763.

\bibitem{blip}
J.~Li, D.~Li, S.~Savarese, and S.~Hoi, ``Blip-2: Bootstrapping language-image pre-training with frozen image encoders and large language models,'' in \emph{International conference on machine learning}.\hskip 1em plus 0.5em minus 0.4em\relax PMLR, 2023, pp. 19\,730--19\,742.

\bibitem{languagebind}
B.~Zhu, B.~Lin, M.~Ning, Y.~Yan, J.~Cui \emph{et~al.}, ``Languagebind: Extending video-language pretraining to n-modality by language-based semantic alignment,'' \emph{arXiv preprint arXiv:2310.01852}, 2023.

\bibitem{uio2}
J.~Lu, C.~Clark, S.~Lee, Z.~Zhang, S.~Khosla, R.~Marten, D.~Hoiem, and A.~Kembhavi, ``Unified-io 2: Scaling autoregressive multimodal models with vision language audio and action,'' in \emph{Proceedings of the IEEE/CVF Conference on Computer Vision and Pattern Recognition}, 2024, pp. 26\,439--26\,455.

\bibitem{scenetrilogy}
P.~N. Chowdhury, A.~K. Bhunia, A.~Sain, S.~Koley, T.~Xiang, and Y.-Z. Song, ``Scenetrilogy: On human scene-sketch and its complementarity with photo and text,'' in \emph{Proceedings of the IEEE/CVF Conference on Computer Vision and Pattern Recognition}, 2023, pp. 10\,972--10\,983.

\bibitem{fashionntm}
A.~Pal, S.~Wadhwa, A.~Jaiswal, X.~Zhang, Y.~Wu, R.~Chada, P.~Natarajan \emph{et~al.}, ``Fashionntm: Multi-turn fashion image retrieval via cascaded memory,'' in \emph{Proceedings of the IEEE/CVF International Conference on Computer Vision}, 2023, pp. 11\,323--11\,334.

\bibitem{GASKN}
Y.~Yu, M.~Liang, M.~Yin, K.~Lu, J.~Du, and Z.~Xue, ``Unsupervised multimodal graph contrastive semantic anchor space dynamic knowledge distillation network for cross-media hash retrieval,'' in \emph{2024 IEEE 40th International Conference on Data Engineering (ICDE)}, 2024, pp. 4699--4708.

\bibitem{mkg}
J.~Zheng, M.~Liang, Y.~Yu, Y.~Li, and Z.~Xue, ``Knowledge graph enhanced multimodal transformer for image-text retrieval,'' in \emph{2024 IEEE 40th International Conference on Data Engineering (ICDE)}, 2024, pp. 70--82.

\end{thebibliography}
}

\end{document}